\pdfoutput=1

\documentclass[11pt]{article}

\usepackage[final]{acl}

\usepackage{times}
\usepackage{latexsym}

\usepackage[T1]{fontenc}

\usepackage[utf8]{inputenc}

\usepackage{microtype}

\usepackage{inconsolata}

\usepackage{graphicx}

\usepackage{microtype}
\usepackage{hyperref}
\usepackage{url}
\usepackage{color}
\usepackage{wrapfig}
\usepackage{natbib}
\usepackage{colortbl}
\definecolor{mygray}{gray}{0.9}
\usepackage{microtype}
\usepackage{graphicx}
\usepackage{booktabs} 
\usepackage{array}
\usepackage[caption=false,font=normalsize,labelfont=sf,textfont=sf]{subfig}
\usepackage{textcomp}
\usepackage{stfloats}
\usepackage{float}
\usepackage{verbatim}
\usepackage[ruled, linesnumbered, lined]{algorithm2e}

\usepackage{multirow}
\usepackage{enumitem}
\usepackage{xcolor}
\usepackage{amsmath,amsfonts}
\definecolor{ggg}{RGB}{26,179,0}
\definecolor{rrr}{RGB}{179,0,0}
\definecolor{oodc}{RGB}{31,73,121}
\definecolor{idc}{RGB}{68,142,68}

\newcommand{\Tmat}[0]{\ensuremath{{\bf T}} }
\newcommand{\Cmat}[0]{\ensuremath{{\bf C}} }

\usepackage{authblk}
\usepackage{amssymb} 

\usepackage{amsmath,amsfonts,bm}

\def\eqref#1{equation~\ref{#1}}

\def\1{\bm{1}}

\def\vx{{\bm{x}}}
\def\vy{{\bm{y}}}

\DeclareMathAlphabet{\mathsfit}{\encodingdefault}{\sfdefault}{m}{sl}
\SetMathAlphabet{\mathsfit}{bold}{\encodingdefault}{\sfdefault}{bx}{n}

\def\gD{{\mathcal{D}}}

\DeclareMathOperator*{\argmin}{arg\,min}

\title{APLOT: Robust Reward Modeling via Adaptive Preference Learning with Optimal Transport}

\author{
 \textbf{Zhuo Li\textsuperscript{1,2,3}},
 \textbf{Yuege Feng\textsuperscript{4}},
 \textbf{Dandan Guo\textsuperscript{5,6}}\thanks{Co-corresponding author, \textit{guodandan@jlu.edu.cn}.},
 \textbf{Jinpeng Hu\textsuperscript{7}},
 \textbf{Anningzhe Gao\textsuperscript{1}}\thanks{Co-corresponding author, \textit{anningzhegao@gmail.com}.},
 \textbf{Xiang Wan\textsuperscript{1}},
 \\
 \textsuperscript{1} Shenzhen International Center for Industrial and Applied Mathematics, \\
 \textsuperscript{2} Shenzhen Research Institute of Big Data, \\
 \textsuperscript{3} The Chinese University of Hong Kong, Shenzhen,
 \textsuperscript{4} Birmingham City University, \\
 \textsuperscript{5} Jilin University,
 \textsuperscript{6} KAUST,
 \textsuperscript{7} Hefei University of Technology,
}

\begin{document}
\maketitle

\begin{abstract}
The reward model (RM) plays a crucial role in aligning Large Language Models (LLMs) with human preferences through Reinforcement Learning, where the Bradley-Terry (BT) objective has been recognized as simple yet powerful, specifically for pairwise preference learning. However, BT-based RMs often struggle to effectively distinguish between similar preference responses, leading to insufficient separation between preferred and non-preferred outputs. Consequently, they may easily overfit easy samples and cannot generalize well to Out-Of-Distribution (OOD) samples, resulting in suboptimal performance. To address these challenges, this paper introduces an effective enhancement to BT-based RMs through an adaptive margin mechanism.  Specifically, we design to dynamically adjust the RM focus on more challenging samples through margins, based on both semantic similarity and model-predicted reward differences, which is approached from a distributional perspective solvable with Optimal Transport (OT). By incorporating these factors into a principled OT cost matrix design, our adaptive margin enables the RM to better capture distributional differences between chosen and rejected responses, yielding significant improvements in performance, convergence speed, and generalization capabilities. Experimental results across multiple benchmarks demonstrate that our method outperforms several existing RM techniques, showcasing enhanced performance in both In-Distribution (ID) and OOD settings. Moreover, RLHF experiments support our practical effectiveness in better aligning LLMs with human preferences. Our code is available at \url{https://github.com/BIRlz/APLOT}.

\end{abstract}

\section{Introduction}
Reinforcement Learning from Human Feedback (RLHF)~\citep{ouyang2022traininglanguagemodelsfollow,rafailov2024direct,deepseekai2025deepseekr1incentivizingreasoningcapability} has emerged as a particularly effective approach in improving the effectiveness and helpfulness of Large Language Models (LLMs)~\citep{openai2024gpt4technicalreport,touvron2023llamaopenefficientfoundation,yang2024qwen2}, and achieving better alignment with human preferences in various fields of artificial intelligence (AI)~\citep{chatgpt,cobbe2021training,shao2024deepseekmathpushinglimitsmathematical,suzgun2022challenging,hu2025agentmentalinteractivemultiagentframework,hu2025beyond,dai2025psyche,hu2023simple,hu2022graphenhancedcontrastivelearning,li2025addoneinincrementalsampleselection,hu2024psycollm}. RLHF begins with optimizing a reward model (RM), which produces feedback that quantifies the quality and correctness of users’ preferences of the provided responses, and thus maximizing reward will direct the LLMs to model effectively satisfy human queries~\citep{ouyang2022traininglanguagemodelsfollow}.

Current RM methods can be broadly categorized into discriminative~\citep{ouyang2022traininglanguagemodelsfollow} and generative approaches~\citep{zheng2023judgingllmasajudgemtbenchchatbot}. Among discriminative methods, a prevalent strategy involves pairwise comparison-based learning, which aims to rank preferred and non-preferred responses based on human annotation by leveraging implicit objectives, such as the Bradley-Terry (BT) model~\citep{bradley1952rank}. While BT model has achieved certain successes, it still faces several limitations, including ``over-optimization''~\citep{gao2023scaling,coste2023reward} that describes a phenomenon where the policy optimization strategy seemingly enhances the proxy reward model but actually leads to the degeneration of the true reward function.

\begin{figure}[t]
    \centering
    \includegraphics[width=1\linewidth]{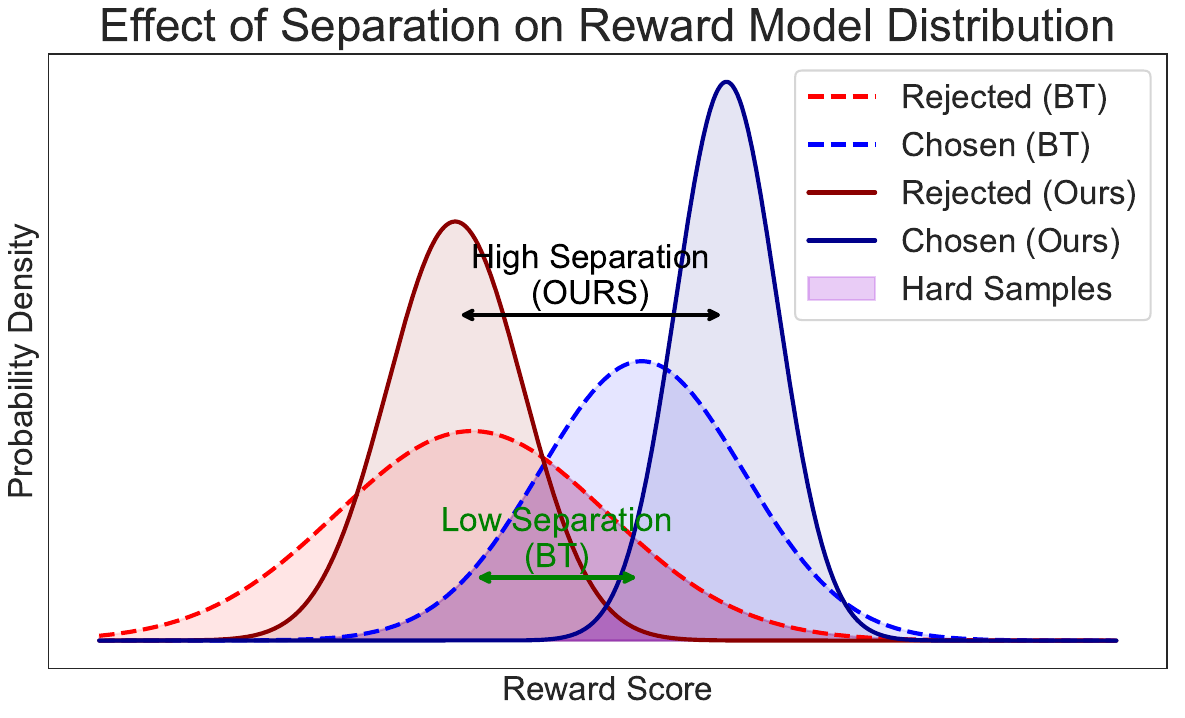}
    \caption{Illustration of the limitation of the traditional BT-based reward model, which only enforces higher scores for chosen samples over rejected ones, neglecting the magnitude of the score difference and resulting in low separation between reward distributions, particularly for hard samples. Our method achieves significantly improved separation, leading to better reward modeling.}
    \label{fig:teaser}\vspace{-1em}
\end{figure}

To address this, several studies have focused on enhancing the reward model with constrained proxy optimization~\citep{dubois2023alpacafarm,yang2024regularizing,chan2024dense,touvron2023llamaopenefficientfoundation} or ensemble techniques~\citep{yang2024rewardsincontextmultiobjectivealignmentfoundation,wang2024interpretablepreferencesmultiobjectivereward,coste2023reward,eisenstein2023helping}. However, these resulting reward models still struggle to distinguish between similar responses, especially when the reward differences are subtle, leading to insufficient separation between preferred and non-preferred responses, resulting in suboptimal model performance and over-fitting to easy samples~\citep{yang2024regularizing,wang2024secretsrlhflargelanguage}. As shown in Figure~\ref{fig:teaser}, BT-based preference learning methods stem from the idea of ranking and only require that chosen samples receive higher scores than rejected samples. This approach neglects the relative magnitude of the score difference, leading to the observed low separation between reward distributions, especially for hard samples. 


In this paper, we introduce an adaptive margin to enhance the pairwise BT reward model that is formulated from a distribution-aware perspective using Optimal Transport (OT)~\citep{cuturi2013sinkhorn}, enabling improved differentiation between preference responses. Our core idea is to dynamically adjust the learning difficult of each training triplet through adaptive margin based on its semantic similarity and model-predicted reward difference. As a result, our proposed method yields a significantly higher separation, as shown in Figure~\ref{fig:teaser}, and enables more effective discrimination between positive and negative examples for improved reward modeling, ensuring that the RM focuses more on challenging samples while avoiding overfitting on easier ones. 

Specifically, we model the margin between the distribution of chosen responses and that of rejected responses as an OT distance, which naturally captures the distributional differences between the two response types. By incorporating both semantic similarity and reward differences into the cost matrix design, OT provides a principled way to estimate desired margins in an adaptive way. Finally, we can incorporate margins into preference learning objective to optimize an improved RM with better performance and robustness in both In-Distribution (ID) and Out-Of-Distribution (OOD) settings. Additional, our approach outperforms several popular RMs across multiple benchmarks, validating its effectiveness and practical utility. Moreover, we observe that our method helps with faster convergence speed without significant additional training consumption. We summarize our contributions as follows:
\begin{enumerate}[leftmargin=*,align=left]
    \item We propose a novel adaptive margin mechanism to improve pairwise reward models, formulated from a distribution-aware perspective using OT. 
    \item Our approach enhances the reward model's ability to better distinguish between similar preference responses by adaptively focusing on challenging samples, by considering both semantic similarity and predicted reward difference.
    \item Experiments show that our method achieves significant improved performance and robustness in both ID and OOD settings on multiple benchmarks, along with faster convergence speed, validating its effectiveness and practical utility.
\end{enumerate}

\section{Related Work}
{\textbf{Reward modeling}} is a critical component in preference learning and RLHF, broadly categorized into discriminative and generative approaches. For the former, classical methods can be traced back to the Bradley-Terry model~\citep{bradley1952rank} and Plackett-Luce model~\citep{plackett1975analysis, luce1959individual}, which optimize an implicit reward function (e.g., a classifier) by learning to imitate human preferences in a pairwise or listwise ranking loss, respectively. Recent studies are more centralized around designing advanced reward models by introducing multi-objective reward functions~\citep{yang2024rewardsincontextmultiobjectivealignmentfoundation,wang2024interpretablepreferencesmultiobjectivereward}, increasing the quality and quantity of training samples~\citep{dubois2023alpacafarm}, regularizing hidden states~\citep{yang2024regularizing}, learning dense rewards~\citep{chan2024dense}, and causal learning~\citep{liu2024rrmrobustrewardmodel}. In addition, several popular preference datasets are proposed to help train a robust reward model, such as the Unified-Feedback (UF) Preference dataset\footnote{\url{https://huggingface.co/datasets/llm-blender/Unified-Feedback}} and the Skywork Preference dataset-80K (SP)~\cite{liu2024skywork}. 

Generative reward modeling is an alternative to classifier-based discriminative reward models by directly employing an LLM to generate a judgment between responses. Generative models excel in providing nuanced, interpretable assessments, capturing subtle differences in language use, and offering deeper insights into the decision-making process~\citep{liu2024skywork}. In addition, some works have emerged to fine-tune models specifically for the task of rating or choosing responses from LLMs~\citep{wang2024directjudgementpreferenceoptimization} and others use the policy LM itself as a generative reward model via prompting it to behave as a judge, in order to achieve better performance of generative reward models~\citep{wang2024self}. Our work is more in line with the discriminative approach.

{\textbf{Adaptive Margin Estimation}} has been explored in various machine learning domains. In metric learning, adaptive margins are used to enforce varying separation distances between classes based on their intrinsic similarity~\citep{sohn2016improved,wu2017sampling}. For instance,~\citet{sohn2016improved} proposed a dynamic margin for triplet loss, where the margin is adjusted based on the difficulty of the triplet. Similarly, in contrastive learning, adaptive margins have been used to improve the discriminative power of learned representations~\citep{khosla2020supervised}. In the context of preference learning, adaptive margins have been less explored but hold significant potential.~\citet{touvron2023llamaopenefficientfoundation} introduces a margin-based regularization term in vanilla BT training objective to help differentiate preferred and non-preferred responses. However, this approach relies on fixed or heuristic margin assignments by requiring additional human annotations, which introduces more consumption and may not fully capture the nuanced differences between responses.

\section{Background}
\textbf{Reinforcement Learning from Human Feedback (RLHF)} serves as a pivotal method for aligning LLMs with human preferences, particularly in terms of their helpfulness and harmlessness \citep{ouyang2022traininglanguagemodelsfollow}. 
RLHF begins with learning a latent reward model $r(\vx,\vy)$ that can implicitly capture human preferences for pairwise comparisons, which are often nuanced or subjective to be explicitly defined \citep{ouyang2022traininglanguagemodelsfollow}. 
Specifically, given a collection of human preference data $\mathcal{D}_\text{p}=\{(\vx,\vy^{w},\vy^{l})\}$, where $\vx$ is a user input prompt, $\vy^{w}, \vy^{l}$ are the preferred (chosen) and non-preferred (rejected) responses, a reward model is usually optimized by minimizing a ranking loss following the Bradley-Terry \citep{bradley1952rank} objective:
\begin{equation}\label{eq:bt-ranking-loss}
     -\mathbb{E}_{(\vx,\vy^{w}, \vy^{l})\sim \gD_\text{p}} \Big[\log \sigma (r(\vx,\vy^{w})-r(\vx,\vy^{l}))\Big],
\end{equation}
where $\sigma(\cdot)$ is the Sigmoid function. Intuitively, \eqref{eq:bt-ranking-loss} induces $r(\vx,\vy)$ to assign a higher reward score to the preferred pairs $(\vx,\vy^{w})$ than the rejected response $(\vx,\vy^{l})$ with respect to an input $\vx$. Therefore, the optimized reward model serves as a proxy for human preferences, enabling the subsequent RL fine-tuning phase. While effective, the vanilla \eqref{eq:bt-ranking-loss} objective also suffers from the lack of sufficient distinction between similar responses \citep{wang2024secretsrlhflargelanguage,touvron2023llamaopenefficientfoundation}, especially when faced with ambiguous training samples. With a learned RM $r(\vx,\vy)$, RLHF optimizes the target LLM policy $\pi_\theta(\vy|\vx)$ for each input $\vx$ by maximizing $\mathbb{E}_{\vx\sim\mathcal{D}, \vy\sim\pi_{\theta}(\vy|\vx)} [r(\vx,\vy) - \text{KL}(\pi_{\theta}(\vy|\vx) \Vert\pi_{\text{ref}}(\vy|\vx))]$. To solve the RLHF objective, Proximal Policy Optimization (PPO) \citep{schulman2017proximalpolicyoptimizationalgorithms} and Group Reward Proxy Optimization (GRPO)~\citep{shao2024deepseekmathpushinglimitsmathematical} have been recognized as the mainstream optimization algorithms \citep{rafailov2024direct}. Recently, several simplified alignment methods have been proposed to avoid the significant generation cost required by online RLHF methods, such as~\citet{rafailov2024direct,du2025simplifyrlhfrewardweightedsft}. Beyond aligning with human preferences, the paradigm of RLHF has also successfully expanded into other NLP tasks, such as prompt refinement~\citep{li-etal-2025-self} and LLM safety detection~\citep{du2024detectingaiflawstargetdriven}.


\textbf{Optimal Transport (OT)} is a popular measurement for comparing distributions \citep{peyre2019computational}, which has been successfully applied on various machine learning tasks~\citep{10904873, NEURIPS2022_a39a9ace, NEURIPS2023_bdabb5d4}. We mainly consider the discrete form in this manuscript. Given two sets of points, their discrete distributions can be formulated as $P=\sum_{n=1}^{N}u_n\delta_{x_n}$ and $Q=\sum_{m=1}^{M} v_m\delta_{y_m}$, where $\delta$ is the Dirac function, and $\boldsymbol{u}$ and $\boldsymbol{v}$ are probability distributions summing to 1. The OT distance between $P$ and $Q$ can be measured as:
\begin{equation}
\min\limits_{\Tmat \in \Pi{(P,Q)}}\langle \Tmat, \Cmat\rangle = \sum_n^{N}\sum_m^{M}T_{nm}C_{nm},
\label{ot}
\end{equation}
where $\Cmat \in \mathbb{R}_{>0}^{n \times m}$ is the cost matrix (e.g., cosine distance) whose each element denotes the distance between $x_n$ and $y_m$, and the transport probability matrix $\Tmat \in \mathbb{R}_{>0}^{N \times M}$ satisfies:
\begin{equation}
\Pi(P,Q):= \left\{\mathbf{T}|\sum_{n=1}^{N}T_{nm}=v_m,\sum_{m=1}^{M}T_{nm}=u_n\right\}.
\end{equation}
As directly optimizing \eqref{ot} can be computationally expensive, an entropic constraint $H(\Tmat) =-\sum_{nm} T_{nm} \ln T_{nm}$ is often introduced for faster optimization \citep{cuturi2013sinkhorn}.

\section{Method}
This section presents our adaptive margin estimation method for robust reward modeling, formulated from a distribution-aware perspective that can be solved by Optimal Transport. Our core idea is to dynamically adjust the margin for each training triplet based on their inherent semantic similarity and model-predicted reward difference, ensuring that the model focuses more on difficult samples while avoiding over-fitting on easy ones.

\textbf{Motivation.} Intuitively, a reasonable margin $\mu_i$ should reflect the difficulty of distinguishing between a preferred $\vy^{w}_{i}$ and non-preferred response $\vy^{l}_{i}$ for an input prompt $\vx_i$. Specifically, the margin $\mu_i$ should be larger for samples with high semantic similarity but low reward difference, indicating that the model finds it challenging to differentiate between them. Conversely, it should be smaller for samples with high reward differences, where the model already demonstrates a clear preference, thereby reducing over-fitting. 

\textbf{Reward Margin Formulation.} Given a set of preference triplets $\{(\vx_i, \vy^{w}_{i}, \vy^{l}_{i})\}_{i=1}^N$, for arbitrary two pairs of preference $(\vx_i, \vy^{w}_{i})$ and $(\vx_j,  \vy^{l}_{j})$, we define the corresponding predicted reward difference as $\Delta f_{ij} = r(\vx_i, \vy^{w}_{i}) - r(\vx_j, \vy^{l}_{j})$, and the semantic similarity between $(\vx_i, \vy^{w}_{i})$ and $(\vx_j, \vy^{l}_{j})$ as $S_{ij} = S\left((\vx_i, \vy^{w}_{i}), (\vx_j, \vy^{l}_{j}) \right)$, where $S(\cdot,\cdot)$ is a measure of semantic similarity for the input (e.g., cosine similarity). To estimate the adaptive margins $\mu$ for these (preferred, non-preferred) pairs, we first build two discrete probability distributions $P$ and $Q$ as follows:
\begin{equation}
    P =  \sum_{i=1}^N \frac{1}{N}\delta_{(\vx_{i}, \vy^{w}_i)}, \quad Q = \sum_{j=1}^N \frac{1}{N} \delta_{(\vx_j, \vy^{l}_j)},
\end{equation}
where $N$ indicates the number of training triples. Therefore, we can define the margins by OT distance:
\begin{equation}\label{eq:ot_matching}
   \text{OT}(P, Q) = \min \limits_{\mathbf{T}\in\Pi(P, Q)} \langle \mathbf{T},\mathbf{C}\rangle - \beta H(\mathbf{T}),
\end{equation}
where $\beta$ is a hyper-parameter for the entropy constraint $H(\mathbf{T)}$ and the $C_{ij}$ measures the distance between $(\vx_i, \vy^{w}_{i})$ and $(\vx_j, \vy^{l}_{j})$. The transport plan $\mathbf{T}$ satisfies $\Pi(P, Q)=\left\{\mathbf{T}\in\mathbb{R}_+^{N\times N}|\sum_{i=1}^{N}T_{ij}=N_j,\notag \right. \left. \sum_{j=1}^{N}T_{ij}=N_i\right\}$. This formulation allows us to capture the distributional differences between preferred and non-preferred responses in a principled manner.

\begin{algorithm*}[t]
\small
\SetKwInOut{Input}{\textbf{Input}}
\SetKwInOut{Output}{\textbf{Output}}
\Input{Preference Dataset $D_p = \{(\vx_i, \vy^{w}_{i}, \vy^{l}_{i})\}_{i=1}^N$, hyper-parameters $\gamma$.}
\Output{Trained reward model $r$}
Initialize a reward model $r$ with parameters $\theta$\;
\While{not converged}{
  Sample a mini-batch of triplets $\{(\vx_i, \vy^{w}_{i}, \vy^{l}_{i})\}_{i=1}^B \sim D_p$\;
  Compute predicted reward differences $\Delta f_{ij} = r(\vx_i, \vy^{w}_{i}) - r(\vx_j, \vy^{l}_{j})$\;
  Calculate semantic similarities $S_{ij} = S\left((\vx_i, \vy^{w}_{i}), (\vx_j, \vy^{l}_{j}) \right)$ using cosine similarity\;
  Construct cost matrix $\mathbf{C}$ by $\mathbf{C}_{ij} = \gamma \cdot S_{ij} + (1-\gamma) \cdot (1 - \sigma(\Delta f_{ij}))$\;
  Build $P$ and $Q$ by $ P =  \sum_{i=1}^N \frac{1}{N}\delta_{(\vx_i, \vy^{w}_i)}\quad, Q = \sum_{j=1}^N \frac{1}{N} \delta_{(\vx_j, \vy^{l}_j)}$\;
  Solve the OT problem to obtain the optimal transport plan $\mathbf{T}^*$ by $ \mathbf{T}^* = \argmin\limits_{\mathbf{T}\in\Pi(P, Q)} \langle \mathbf{T}, \mathbf{C} \rangle - 0.1 \times H(\mathbf{T})$\;
  Estimate adaptive margins $\mu_i = \sum_{j=1}^B T^*_{ij} \cdot C_{ij}$ for each triplet in the mini-batch\;
  Compute the adjusted Ranking Loss $\mathcal{L}$ using the estimated adaptive margins: $\mathcal{L} = - \frac{1}{B} \sum_{i=1}^B \log \sigma \left( r(\vx_i, \vy^{w}_{i}) - r(\vx_i, \vy^{l}_{i}) - \mu_i \right) $\;
  Update the reward model parameters $\theta$ using gradient descent with the computed loss $\mathcal{L}$\;
}
\caption{Reward Modeling with Adaptive Margin Estimation in Mini-Batch.}
\label{alg:training}
\end{algorithm*}
\textbf{Cost Matrix Design.} Cost matrix acts as a determinant in the optimization of the transport plan between $P$ and $Q$. Although it is possible to use point-wise distances like cosine metric, these only focus on semantic similarity and ignore the reward differences, leading to suboptimal margin estimation. Recall our motivation that the margin should reflect both the semantic similarity and the model's predicted reward differences to capture the true difficulty of distinguishing between preferred and non-preferred responses. Therefore, we design the cost matrix $\mathbf{C}$ to incorporate both semantic similarity $S\left((\vx_i, \vy^{w}_{i}), (\vx_j, \vy^{l}_{j}) \right)$ and reward differences $\Delta f_i= r(\vx_i, \vy^{w}_{i}) - r(\vx_j, \vy^{l}_{j})$ by:
\begin{equation}
C_{ij} =  \underbrace{\gamma \cdot S_{i,j}}_{\text{Semantic Similarity}} + \quad \underbrace{(1-\gamma) \cdot (1-\sigma(\Delta f_{ij})}_{\text{Reward Differences}}
\label{eq:compute_margin}
\end{equation}
where $\gamma$ is a balance hyper-parameter and $\sigma$ is Sigmoid function. We use cosine similarity for $S(\cdot,\cdot)$. Clearly, larger semantic similarity leads to a significant increase in cost, while larger reward differences only bring a slight improvement, aligning with our motivation. As a result, this formulation adaptively captures the learning differences between preferred and non-preferred responses. For semantic measurement, we extract the last hidden state of the last non-pad token in each input pair.

\textbf{Distributional Adaptive Margin Estimation and Training Loss.} Using this carefully designed cost matrix $\mathbf{C}$, we can compute the optimal transport plan $\mathbf{T}^*$ by \eqref{eq:ot_matching}. As a result, the adaptive margin $\mu_i$ for the i-th triplet is then derived from the $\mathbf{T}^*$ and $\mathbf{C}$, which we name ``APLOT'':
\begin{equation}\label{eq:aplot_margin}
    \mu_i = \sum_{j=1}^N {T}^*_{ij}\cdot C_{ij}.
\end{equation}
This formulation ensures that the margin for each triplet is influenced by its pairwise relationships with other triplets, as captured by the transport plan. Triplets that are more challenging to distinguish will receive larger margins, while easier triplets will receive smaller margins. Finally, we can incorporate our adaptive margin to BT ranking loss~\eqref{eq:bt-ranking-loss} to optimize a robust reward model by minimizing the following objective:
\begin{equation}\label{eq:final-loss}
 -\mathbb{E}_{(\vx,\vy^{w}, \vy^{l})\sim \gD_\text{p}} \Big[\log \sigma (r(\vx,\vy^{w})-r(\vx,\vy^{l}) - \mu)\Big].
\end{equation}
In addition, we introduce a simpler baseline of our method by estimating the margin $\mu_i$ in a point-to-point way, which we name ``PointMargin'':
\begin{equation}\label{eq:point_margin}
    \mu_i = \sum_{j=1}^NC_{ij}.
\end{equation}
The main difference between APLOT (\eqref{eq:aplot_margin}) and PointMargin (\eqref{eq:point_margin}) lies in the approach to aggregating cost information contained in the matrix $\mathbf{C}$, where PointMargin computes the margin as a simple summation of costs, providing a straightforward but local, point-to-point baseline with less training consumption. APLOT, in contrast, takes a more sophisticated distributional approach, leveraging the optimal transport plan $\Tmat$ to derive a margin that is a weighted sum of costs, thereby reflecting a globally optimal matching process across the distributions of preferred and dispreferred responses with additional computation.

\textbf{Why Margin can help RM better differentiate the responses?} The introduction of a margin $\mu$ enhances the RM's ability to differentiate between preferred and dispreferred responses given an input $\vx$. Minimizing~\eqref{eq:final-loss} encourages the score difference $ = r(\vx, \vy^{w}) - r(\vx, \vy^{l})$ to exceed $\mu$, thereby enforcing a more pronounced separation between preferred and dispreferred responses. When $s \leq \mu$, the model receives stronger gradient signals (e.g., gradient magnitude $1 - (s - \mu) \geq 0.5$), guiding it to focus on harder samples with small but positive margins. A larger margin $\mu$ naturally correlates with higher difficulty, prompting the RM to concentrate on finer-grained distinctions and improve learning where it matters most. Our method further benefits from adaptively estimating $\mu$, allowing the learning objective to dynamically adjust based on the difficulty of sample pairs. More gradient analysis is provided in Appendix~\ref{app:margin-analysis}.


In summary, by formulating the adaptive margin estimation as an Optimal Transport problem, we gain a principled way to incorporate both semantic similarity and predicted reward differences into the margin. This distribution-aware method adjusts $\mu_i$ based on the specific characteristics of each triplet, providing a robust defense against over-fitting while directing the model's attention towards the most challenging cases. This approach ultimately improves the model's performance in reward modeling tasks by ensuring consistent and robust reward assignments. We summarize the training algorithm in Alg.~\ref{alg:training}.

\begin{table*}[!htbp]
\centering
\caption{\small{Results on \textcolor{idc}{ID} and \textcolor{oodc}{OOD} evaluation with 40K training data from Unified Feedback based on gemma-2b-it. The best performance in each task is \textbf{bold}. Baseline results are cited from~\citet{yang2024regularizing}. We set HardMargin as 1.0.}}
\resizebox{0.99\textwidth}{!}{
\begin{tabular}{c|ccc|cccccc}
\toprule
\multirow{2}{*}{Method} & \multicolumn{1}{c}{\textcolor{idc}{Unified}} & \multicolumn{1}{c}{\textcolor{oodc}{HHH}} & \multicolumn{1}{c}{\textcolor{oodc}{MT}} & \multicolumn{5}{c}{\textcolor{oodc}{RewardBench}} \\
 & \textcolor{idc}{Feedback} & \textcolor{oodc}{Alignment} & \textcolor{oodc}{Bench} &  \textcolor{oodc}{Avg.} & \textcolor{oodc}{Chat} & \textcolor{oodc}{Chat-Hard} & \textcolor{oodc}{Safety} & \textcolor{oodc}{Reasoning} \\  \midrule \midrule
BT - Vanilla & 68.8 & 70.3 & 69.1 & 64.5 & 95.8 & 37.3 & 59.9 & 64.8 \\
BT - HardMargin & 69.6 &  69.8 &  71.0 & 66.1 & 97.2 & 37.5 & 56.8 & 72.7 \\
BT - LabelSmooth & 68.5  & 68.8 &  71.9 & 61.1 & 91.6 & 39.0 & 53.8 & 60.2 \\
BT - Ensemble &  69.9  & 72.2 &  71.1 & 65.2 & 96.1 & 38.2 & 58.8 & 67.6 \\ 
GRM + DPO & 70.2  & 71.6 &  71.3 & 70.8 & {97.8} & 42.1 & 77.9 & 65.2 \\ 
GRM + DPO-NoRef & 71.4  & 76.6 &  72.1 & 66.6 & {92.5} & 39.9 & 72.5 & 61.4 \\ 
GRM + SFT & 71.5  & 78.7 &  73.0 & 66.8 & {94.1} & 41.9 & 69.5 & 61.5 \\ \midrule
APLOT (OURS) &\textbf{73.84} & \textbf{81.25} & \textbf{75.23} & \textbf{74.10} & \textbf{97.21} & \textbf{42.54} & \textbf{80.81} & \textbf{72.17} \\ \bottomrule
\end{tabular}
}\label{res_main_1}
\end{table*}

\section{Experiment}
\subsection{Setup}
\paragraph{\textbf{Datasets}.} By following~\citet{yang2024regularizing}, we leverage the Unified-Feedback (UF) preference dataset\footnote{\url{https://huggingface.co/datasets/llm-blender/Unified-Feedback}} to demonstrate the effectiveness of our method when compared with vanilla BT training objective. For a comprehensive evaluation of the reward model's performance, we consider both in-distribution (ID) and out-of-distribution (OOD). Specifically, not only are models evaluated on the standard 1K UF eval set (ID), but we also compare the performance of different RM methods on three popular OOD datasets: HHH-Alignment~\citep{coste2023reward}, MT-Bench Human Judgments~\citep{zheng2023judgingllmasajudgemtbenchchatbot}, RewardBench~\citep{lambert2403rewardbench} and RM-Bench~\citep{liu2024rmbenchbenchmarkingrewardmodels}. The HHH-Alignment dataset mainly evaluates a language model from the perspectives of helpfulness, honesty, and harmlessness. MT-Bench contains 3.3K expert-level pairwise human preferences for model responses generated by 6 models in response to MT-bench questions. Besides, RewardBench is a popular benchmark designed to comprehensively evaluate the capabilities and safety of reward models. On the other hand, we also adopt the Skywork Reward Preference (SP) dataset~\citep{liu2024skywork} for reward model training when compared with several SOTA RM methods.

\paragraph{\textbf{Base Models} and \textbf{Training Details}.} For base models, we adopt gemma-2b-it~\citep{team2024gemma} and Llama-3.1-8B-Instruct~\citep{dubey2024llama}. Training details can be found in Appendix~\ref{app:train_de}.


\paragraph{\textbf{Baselines}.} We evaluate the performance of our method with several baselines, including (1) Vanilla BT reward model; (2) BT-Variants, including BT w/ margin, label smooth, PosReg and ensemble~\citep{touvron2023llamaopenefficientfoundation,wang2024secretsrlhflargelanguage}; (4) GRM~\citep{yang2024regularizing} that designs to regularize the hidden states by incorporating a DPO loss~\citep{rafailov2024direct} and its variants. 

\subsection{Results and Analysis}
\paragraph{\textbf{Performance on ID and OOD settings compared with baselines}.} By following~\citet{yang2024regularizing}, we first randomly sample 40K samples from the UF dataset for RM training with the help of LoRA~\citep{hu2021loralowrankadaptationlarge}. As shown in Table~\ref{res_main_1}, our method consistently outperforms all baseline methods on both ID and OOD evaluation tasks. Specifically, our method achieves the highest scores of {73.84} on Unified Feedback, {81.25} on HHH Alignment, and {75.23} on MT Bench, indicating that our method significantly improves the model's ability to generalize to both ID and OOD reward evaluation datasets. The strong performance on OOD tasks, in particular, demonstrates the robustness of our approach in handling unseen scenarios. Compared to the baseline methods, our approach shows a clear advantage, especially in scenarios where the model needs to differentiate between similar responses with subtle reward differences. 

\begin{figure}[t]\vspace{-1em}
    \centering
    \includegraphics[width=1\linewidth]{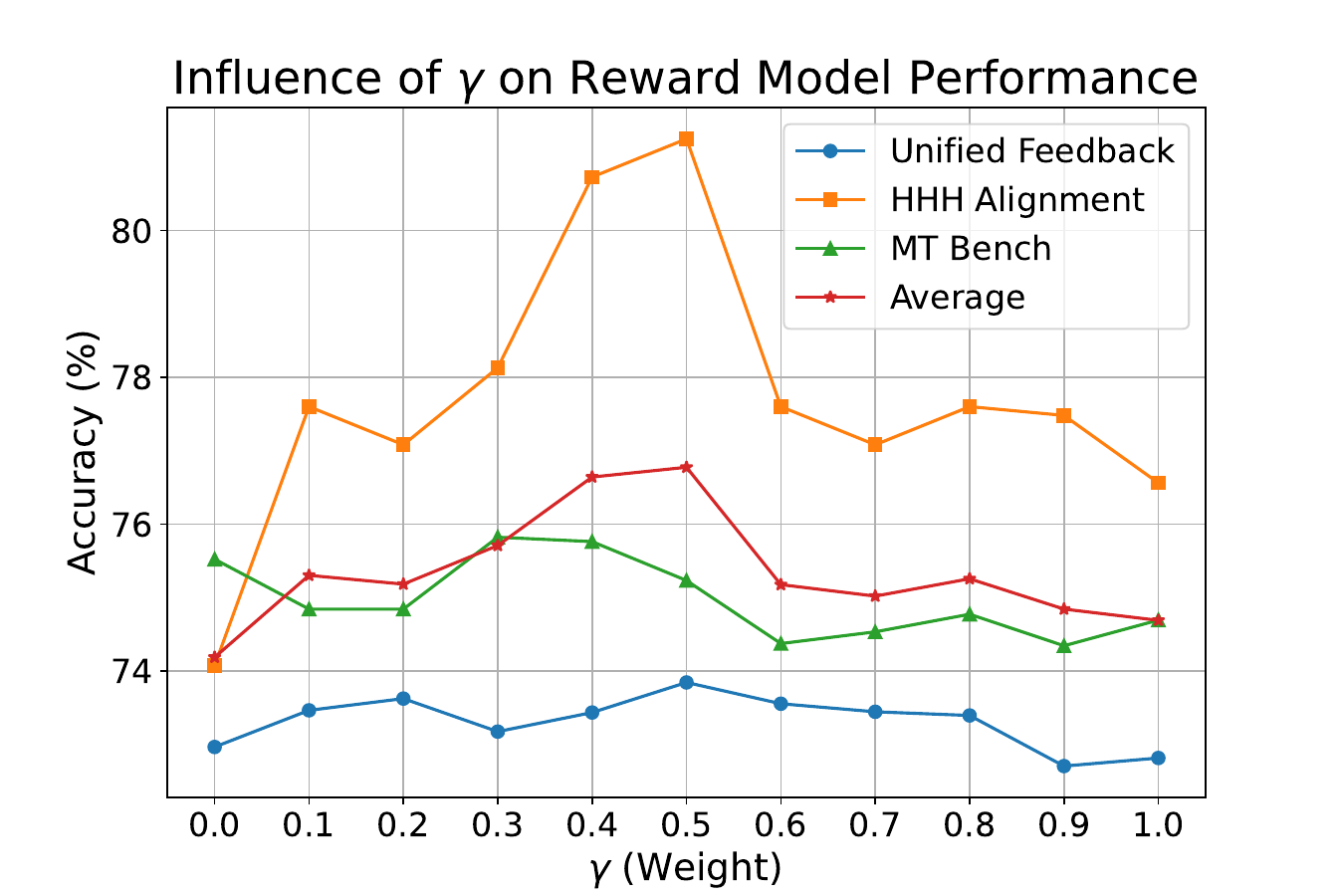}
    \caption{Influence of the weight $\gamma$ on our reward model's performance across different tasks, which balances the semantic consistency and reward difference.}
    \label{fig:ablation_gamma}\vspace{-1em}
\end{figure}
\paragraph{\textbf{Influence Analysis of $\gamma$.}} The hyperparameter $\gamma$ balances the importance of semantic similarity and reward difference in the cost matrix. To quantitatively evaluate the influence of $\gamma$, we train RMs on the a 40K subset of UF dataset based on gemma-2b-it equipped with Lora. As shown in Figure~\ref{fig:ablation_gamma}, optimal performance across tasks is achieved when $\gamma$ is around 0.4 to 0.6, with $\gamma=0.5$ consistently leading to the highest accuracy in three test datasets, as well as the overall average performance. When $\gamma$ is too high or too low, the model's performance decreases, as it causes the model to disproportionately emphasize one aspect over the other. This indicates that a balanced $\gamma$ is crucial for our approach to effectively incorporate both semantic similarity and reward difference into margin estimation, thereby enhancing the reward model's performance. Our method performs best when $\gamma$ is set around 0.5, highlighting the importance of balancing these two factors in our cost matrix design, specifically for OOD settings.

\begin{table*}[!t]
\centering
\caption{\small{Performance comparison of different reward models on RewardBench. The best performance in each task is in \textbf{bold} and we cite results from~\citet{liu2024skywork}.}}
\resizebox{0.8\textwidth}{!}{
\begin{tabular}{c|lccccc}
\toprule
Type & Method & Avg. & Chat & Chat-Hard & Safety & Reasoning \\ \midrule \midrule
\multirow{4}{*}{\rotatebox[origin=c]{90}{Generative}}&SFR-LLaMa-3.1-70B-Judge-I  & 92.7 & 96.9 & 84.8 & 91.6 & \textbf{97.6} \\ 
&Gemini-1.5  & 86.8 & 94.1 & 77.0 & 85.8 & 90.2 \\ 
&GPT-4o  & 86.7 & 96.1 & 76.1 & 88.1 & 86.6 \\ 
&SFR-nemo-12B-Judge-r  & 90.3 & 97.2 & 82.2 & 86.5 & 95.1 \\ \midrule
\multirow{8}{*}{\rotatebox[origin=c]{90}{Discriminative}}&Nemotron-340B-Reward  & 92.2 & 95.8 & 87.1 & 92.2 & 93.6 \\ 
&ArmoRM-Llama3-8B-v0.1  & 90.8 & 96.9 & 76.8 & 92.2 & 97.3 \\ 
&InternLM-20B-Reward  & 90.2 & \textbf{98.9} & 76.5 & 89.9 & 95.8 \\ 
&Llama-3-OffsetBias-RM-8B & 89.4 & 97.2 & 81.8 & 86.8 & 91.9 \\ 
&Llama-3.1-BT-RM-8B   & 91.8 & 94.6 & 84.5 & 91.5 & 96.5 \\ 
&Skywork-Reward-Llama-3.1-8B  & 92.5 & 95.8 & 87.3 & 90.6 & 96.2 \\ \cmidrule{2-7}
&\textbf{APLOT-Scratch-Llama-3.1-8B}   & 92.1 & 97.2 & 84.9 & 92.1 & 94.2  \\
&\textbf{APLOT-Skywork-Llama-3.1-8B}   & \textbf{94.4} & 93.9 & \textbf{89.0} & \textbf{93.2} & 97.4 \\ 
\bottomrule
\end{tabular}\vspace{-1em}
}\label{res_rewardbench}
\end{table*}

\begin{table*}[!t]
\centering
\caption{\small Performance comparison of different reward models on RM-Bench. The best performance in each task is in \textbf{bold} and we cite results from~\citet{liu2024rmbenchbenchmarkingrewardmodels}.}
\label{res_RMbench}
\resizebox{0.8\textwidth}{!}{
\begin{tabular}{c|lccccc}
\toprule
{Type} & {Method} & {Avg.} & {Chat} & {Math} & {Code} & {Safety} \\
\midrule \midrule
\multirow{2}{*}{\rotatebox[origin=c]{90}{DPO}} 
& upstage/SOLAR-10.7B-Instruct-v1.0 & 64.8 & \textbf{78.6} & 52.3 & 49.6 & 78.9 \\
& allenai/tulu-2-dpo-13b & 63.8 & 66.4 & 51.4 & 51.8 & 85.4 \\
\midrule
\multirow{8}{*}{\rotatebox[origin=c]{90}{Discriminative}} 
& URM-LLaMa-3.1-8B & 70.0 & 71.2 & 61.8 & 54.1 & 93.1 \\
& Nemotron-340B-Reward & 69.5 & 71.2 & 59.8 & 59.4 & 87.5 \\
& Llama-3-OffsetBias-RM-8B & 69.0 & 71.3 & 61.9 & 53.2 & 89.6 \\
& internlm2-20b-reward & 68.3 & 63.1 & \textbf{66.8} & 56.7 & 86.5 \\
& GRM-llama-3-8B-sftreg & 68.2 & 62.7 & 62.5 & 57.8 & 90.0 \\ 
& Skywork-Reward-Llama-3.1-8B & 70.1 & 69.5 & 60.6 & \textbf{54.5} & 95.7 \\\cmidrule{2-7}
& \textbf{APLOT-Scratch-Llama-3.1-8B} & 71.68 & 72.44 & 63.58 & 54.19 & {96.32} \\
& \textbf{APLOT-Skywork-Llama-3.1-8B} & \textbf{72.10} & {72.82} & {63.89} & {54.24} & \textbf{96.50} \\
\bottomrule
\end{tabular}
}
\end{table*}

\paragraph{\textbf{Performance on RewardBench and RM-Bench}.} Table~\ref{res_rewardbench} presents a comprehensive evaluation of various reward models on the RewardBench dataset, demonstrating the effectiveness of our proposed method. We train our RM on SP with full parameter tuning. Notably, our method achieves compelling results even when trained from scratch. Specifically, ``APLOT-Scratch-Llama-3.1-8B'' (trained scratch from Llama-3.1-8B-Instruct) attains a strong average score of 92.1, comparable to other high-performing reward models and even better than several RM with higher scale, highlighting the inherent effectiveness of our approach. Furthermore, our method exhibits remarkable flexibility by also serving as a powerful tool to enhance pre-existing reward models. By applying our technique to refine the already strong Skywork-Reward-Llama-3.1-8B model, we achieve a further performance boost, reaching a best score of 94.4 with ``APLOT-Skywork-Llama-3.1-8B'' and significantly improving results in key areas like ``Chat-Hard'' (89.0) and ``Safety'' (93.2). While generative models like SFR-LLaMa-3.1-70B-Judge-I excel in specific tasks such as ``Reasoning'' (97.6), our discriminative approach demonstrates both intrinsic efficacy and the ability to enhance other models, showcasing its versatility. In conclusion, these findings underscore the dual advantage of our reward modeling method: strong performance on its own and the capacity to effectively ``plug-and-play'' to elevate the performance of existing reward models. Besides, we also evaluate the performance of our methods on RM-Bench, which is another popular and more challenging RM benchmark. As shown in Table~\ref{res_RMbench}, we also observe that our method brings better performance compared with several strong baselines.

\begin{figure}[t]\vspace{-1em}
    \centering
    \includegraphics[width=1\linewidth]{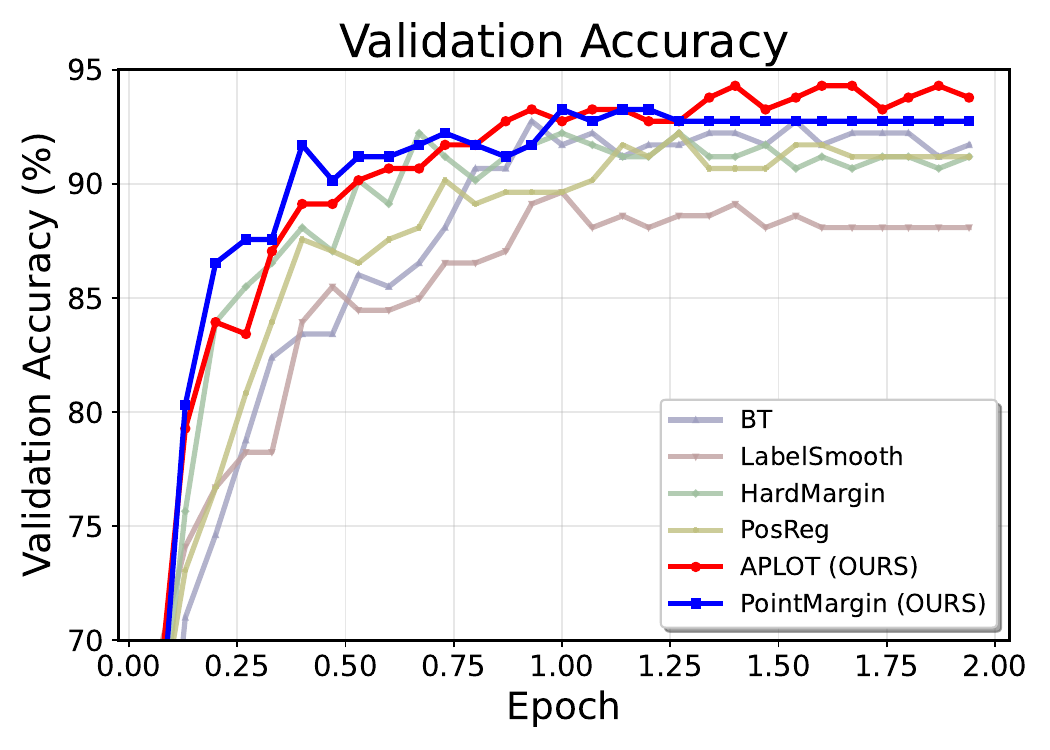}
    \caption{Illustration of the convergence and performance of our proposed method, in terms of validation accuracy over training epochs. Both APLOT and PointMargin demonstrate faster convergence, achieving higher accuracy with fewer epochs, and ultimately reach better accuracies compared to baselines.}
    \label{fig:val_vs_epoch}\vspace{-1em}
\end{figure}

\paragraph{\textbf{Convergence and Validation Accuracy Comparison}.} We evaluate the convergence speed compared with baseline methods. Experimentally, we randomly sampled 40K training points from SP dataset. Figure~\ref{fig:val_vs_epoch} depicting validation accuracy against training epoch, clearly demonstrates the superior convergence and performance of our proposed margin estimators APLOT (\eqref{eq:aplot_margin}) and PointMargin (\eqref{eq:point_margin}).
Experimentally, both APLOT (red line with circles) and PointMargin (blue line with squares) exhibit significant convergence speed learning and better performance by achieving >90\% validation accuracy before 0.5 epoch, compared to baselines that struggle to reach similar accuracy even after 1.0 epoch. This indicates that our methods learn more efficiently and require less training to reach optimal performance. In terms of final performance, PointMargin shows substantial improvement over baselines, while APLOT further enhances performance, surpassing 94\% accuracy after nearly 2.0 epochs, where baselines plateau in the 88\%-92\% range.  In conclusion, these results underscore the effectiveness of our proposed techniques, with PointMargin providing a strong improvement and APLOT's enhanced performance demonstrating the value of a distributional perspective in learning more robust reward models. 

\begin{table}[!htbp]
\centering
\caption{\small{Performance comparison of RMs on the Unified Feedback, HHH Alignment, and MT Bench datasets under 20\% label noise within the training dataset SP.}}
\resizebox{0.45\textwidth}{!}{
\begin{tabular}{c|cccc}
\toprule
\multirow{2}{*}{Method} & \multicolumn{1}{c}{\textcolor{idc}{Unified}} & \multicolumn{1}{c}{\textcolor{oodc}{HHH}} & \multicolumn{1}{c}{\textcolor{oodc}{MT}}   & \multirow{2}{*}{Avg.}  \\
 & \textcolor{idc}{Feedback} & \textcolor{oodc}{Alignment} & \textcolor{oodc}{Bench}  &   \\  \midrule \midrule
BT - Vanilla & 71.53 & 72.92& 72.68  &  72.38\\
BT - HardMargin & 70.43 & 77.60 & 73.85   &  73.96 \\
BT - LabelSmooth  & 70.62 & 76.56& 71.28   &  72.82\\
BT - PosReg & 71.08 & 79.17 &  \textbf{75.11} &    75.21 \\ \midrule
APLOT (OURS)  & \textbf{71.82} & \textbf{81.78}  &  74.88  & \textbf{76.16} \\ \bottomrule
\end{tabular}
}\label{res_noise}
\end{table}

\paragraph{\textbf{Performance against Label-Noise}} Label noise are inevitable during human preference annotations~\cite{wang2024secretsrlhflargelanguage}, which hinders the generalization and effectiveness of reward models. To evaluate the robustness of our method, we design to randomly assign 20\% label noise into a 20K SP training subset. As shown in Table~\ref{res_noise}, we find that several variants bring improvements to the vanilla BT RM, while our method can significantly enhance the reward model by more accurately judging the sample quality, even within the noisy training annotations.

\subsection{Evaluation on RLHF}

\begin{figure*}[t]
    \centering
    \includegraphics[width=1\linewidth]{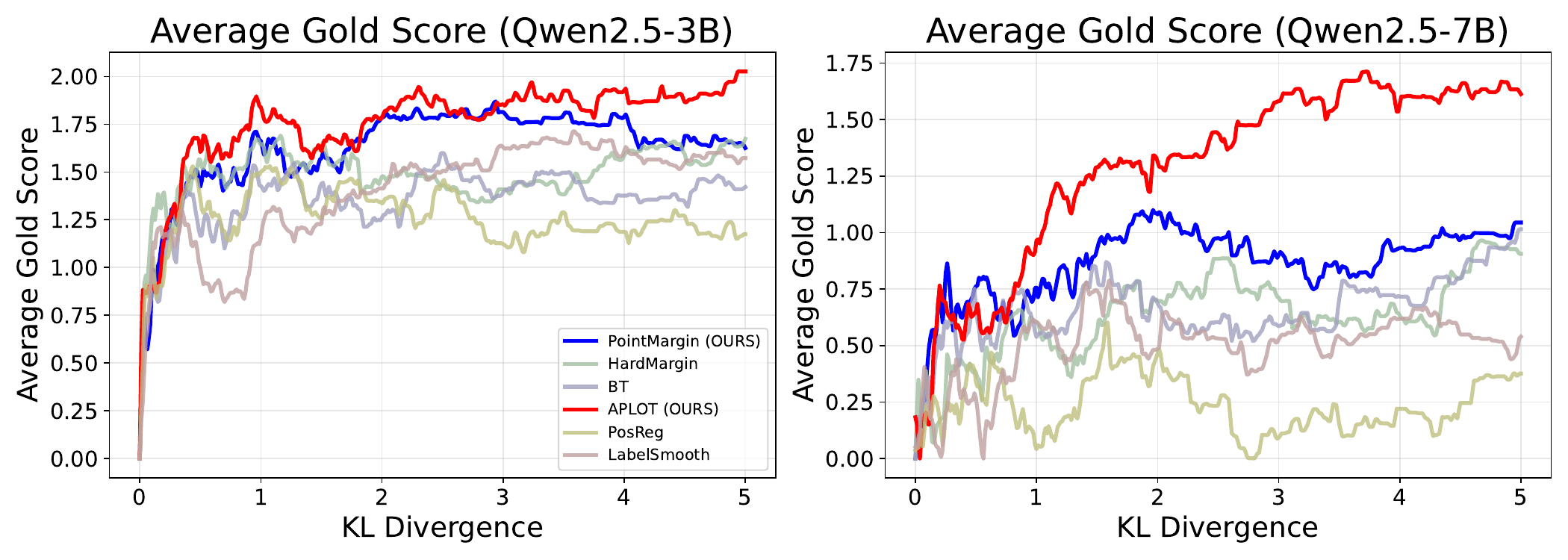}
    \caption{Gold scores from the Best-of-$N$ (Bo$N$) test, using responses sampled from Qwen2.5-3B and 7B-Instruct, respectively. Rewards are normalized to start at 0. APLOT shows robust alignment with gold rewards despite increasing KL Divergence.}
    \label{fig:average_gold_score}\vspace{-1em}
\end{figure*}

\paragraph{\textbf{Best-of-$N$ (Bo$N$) Sampling Test.}} Figure \ref{fig:average_gold_score} reports Bo$N$ results for the Qwen 2.5-3 B and 7 B-Instruct models. For each setting, we train proxy reward models on a randomly sampled 40 K subset of the SP corpus using Llama-3.1-8B-Instruct as the backbone. Following \citet{coste2024rewardmodelensembleshelp, gao2023scaling}, we generate N candidate completions for every prompt in a 300-instance out-of-distribution (OOD) test set, rank them with the proxy RMs, and then evaluate the chosen responses with a high-fidelity gold reward model (Skywork-Reward-Gemma-2-27 B). The average gold score over the 300 prompts thus reflects the true quality of the proxy-selected answers. We vary the KL-divergence budget from 0 to 5, which—through the relation $\text{KL}_{\text{Bo}N}$ = $\text{log}\ N - \frac{N-1}{N}$~\cite{gao2023scaling}—corresponds to N ranging from 1 to 405. A well-behaved proxy RM should yield monotonically increasing proxy and gold scores as the KL budget (and hence N) grows. Several baselines instead plateau or even decline once KL > 2 (see the right-hand plot for Qwen 2.5-7B-Instruct), revealing over-optimization. In contrast, our APLOT approach maintains a steady rise in gold score across the entire KL range, demonstrating its ability to curb over-optimization and underscoring APLOT’s robustness as a proxy reward model for RLHF.

\paragraph{\textbf{PPO.}} Beyond the Bo$N$ Test, we conduct the PPO experiments to practically investigate whether our reward model helps better RLHF training with the help of an adaptive margin. As shown in Table~\ref{res_ppo_rlhf}, our experimental results highlight the better performance of the policy model fine-tuned with our APLOT. On the OpenRLHF-Llama3-8B-SFT base, our APLOT model achieve an average of 58.55, compared to 55.68 for the baseline and 56.94 for SKRM. These findings validate the effectiveness of our APLOT reward model in consistently enhancing the capabilities of language models through PPO fine-tuning. Training and evaluation details can be found in Appendix~\ref{app:ppo_train_eval_de}.

\begin{table}[!htbp]
\centering
\caption{\small{Benchmark Evaluation. Baseline indicates the benchmark performance on vanilla OpenRLHF-Llama3-8B-SFT, respectively. SWRM is that of the policy model trained based on the reward model Skywork-RM-Llama3.1-8B-Instruct, and OURS is that based on our APLOT-RM-Llama3.1-8B-Instruct.}}
\resizebox{0.48\textwidth}{!}{
\begin{tabular}{l|ccc}
\toprule
Benchmark & Baseline & SKRM  & APLOT \\ \midrule \midrule
GSM8K$_{\text{acc}}$ &  74.83 & 78.17   & \textbf{79.23} \\
Hellaswag$_{\text{acc}}$ & 72.51 & 74.76 & \textbf{76.12} \\
IFeval$_{\text{acc}}$ &  44.92 & 45.10 &  \textbf{48.98} \\
MMLU$_{\text{acc}}$ & 54.45 & 52.40   & \textbf{55.62} \\
ProcessBench$_{\text{acc}}$ & 4.46 & 10.31   & \textbf{10.62} \\
Race$_{\text{acc}}$ & 79.21 & 78.82   & \textbf{80.93} \\
BBH$_{\text{acc}}$  & 61.20 & \textbf{62.68}   & {62.42} \\
Humaneval$_{\text{pass@1}}$ &  60.98 & 57.32   & \textbf{63.41} \\
TriviaQA$_{\text{acc}}$ &  48.53 & \textbf{52.86}   & 49.63 \\ \midrule
Avg.  & 55.68 & 56.94 &    \textbf{58.55} \\ \bottomrule
\end{tabular}
}\label{res_ppo_rlhf}
\end{table}

\section{Conclusion}
In this work, we proposed to enhance the pairwise preference reward model with a novel adaptive margin to achieve better generalization in RLHF. Our approach leverages Optimal Transport (OT) to dynamically adjust the margin based on semantic similarity and model-predicted reward difference, ensuring that the model focuses more on challenging samples while avoiding over-fitting on easier ones. Through extensive experiments, we have demonstrated that our method consistently outperforms existing reward modeling techniques across multiple benchmarks, showing significant improvements in both in-distribution and out-of-distribution tasks. The ablation study further highlights the importance of balancing semantic similarity and reward difference in the cost matrix design. 

\section{Acknowledgments}
This work was supported by the Guangxi Key R\&D Project  (No. AB24010167), the Project (No. 20232ABC03A25), and the Futian Healthcare Research Project (No.FTWS002). This work was also supported by the National Natural Science Foundation of China (NSFC) under Grant 62306125 and Grant 62402158.

\section{Limitation and Future Work}
Despite the promising results demonstrated by our method, there are several limitations that need to be acknowledged. Firstly, our current method is designed for language-based reward models and has not been adapted for multi-modal inputs or progress reward models, and cannot be explicitly adapted to generative reward modeling. The complexity of multi-modal data and the dynamic nature of progress reward modeling pose additional challenges that our current approach does not address. Secondly, our method relies on the quality and representativeness of the training data. If the training data is biased or lacks diversity, it may limit the performance of our method.

In future work, we plan to address the limitations mentioned above. We will explore the application of our method in multi-modal reward modeling, where the reward function needs to consider both text and other modalities such as images or audio. This extension will require the development of new techniques to effectively integrate multi-modal information into the reward model. Additionally, we aim to investigate the potential of our method in progress and generative reward modeling, where the reward function needs to adapt to the progress of the learning process. This will involve designing new algorithms that can dynamically adjust the reward function based on the agent's progress. 

\bibliography{acl_latex}

\clearpage
\appendix
\onecolumn

\section{Analysis of Margin $\mu$'s Impact on Loss and Gradient}\label{app:margin-analysis}

Let the per-sample loss for a triplet $(\vx, \vy^{w}, \vy^{l})$ be defined as:
\begin{equation}
    \ell = -\log(\sigma(s - \mu)), \quad s = r(\vx, \vy^{w}) - r(\vx, \vy^{l}), \quad \mu > 0
\end{equation}
where $r(\vx, \vy)$ denotes the model score for input-output pair $(\vx, \vy)$.

\paragraph{Impact on Loss Value.}
The sigmoid function $\sigma(x) = \frac{1}{1 + e^{-x}}$ is monotonically increasing. The negative logarithm $-\log(\cdot)$ is monotonically decreasing. 

\textbf{1. If $s \leq \mu$ (Difficult/Marginal Sample)}, we will have $\sigma(s - \mu) \leq \sigma(0) = 0.5$ and $\ell = -\log(\sigma(s - \mu)) \geq -\log(0.5) = \log 2$. We can find that the loss is substantial, indicating that the model needs to increase the separation between preferred and non-preferred responses. As $s - \mu \to -\infty$, $\sigma(s - \mu) \to 0, \quad \ell \to \infty$.

\textbf{2. If $s > \mu$ (Easy Sample, margin satisfied)}, we will have $\sigma(s - \mu) > 0.5$ and $\ell = -\log(\sigma(s - \mu)) < \log 2$. We observe that as $s - \mu \to \infty$, $\sigma(s - \mu) \to 1, \quad \ell \to 0$. This shows that the loss function penalizes more heavily when the score difference $s$ does not exceed the margin $\mu$.

\paragraph{Impact on Gradient.}
We are also interested in the gradient of the loss $\ell$ with respect to the score difference $s$, which dictates how the model scores are updated. Let $X = s - \mu$, then we can obtain the following derivation:
\begin{equation}
\frac{\partial \ell}{\partial X} = \frac{d}{dX} \left( -\log(\sigma(X)) \right) = -\frac{1}{\sigma(X)} \cdot \sigma'(X)
\end{equation}

Since $\sigma'(X) = \sigma(X)(1 - \sigma(X)), \quad \text{then } \frac{\partial \ell}{\partial X} = -(1 - \sigma(X)) = \sigma(X) - 1$. By chain rule:
\begin{equation}
\frac{\partial \ell}{\partial s} = \frac{\partial \ell}{\partial X} \cdot \frac{\partial X}{\partial s} = \sigma(s - \mu) - 1
\end{equation}

In gradient descent, model parameters are updated in the direction of the negative gradient. Therefore, the effective update signal for increasing $s$ is proportional to:
\begin{equation}
-\frac{\partial \ell}{\partial s} = 1 - \sigma(s - \mu)
\end{equation}

\textbf{1. If $s \leq \mu$ (Difficult/Marginal Sample)}, we will have $\sigma(s - \mu) < 0.5, \quad 1 - \sigma(s - \mu) > 0.5$. As $s - \mu \to -\infty$, $\sigma(s - \mu) \to 0, \quad 1 - \sigma(s - \mu) \to 1$, we obtain a strong gradient signal, encouraging the model to increase $s$ for hard pairs.

\textbf{2. If $s > \mu$ (Easy Sample)}, we will have $\sigma(s - \mu) > 0.5, \quad 1 - \sigma(s - \mu) < 0.5$. As $s - \mu \to \infty$, $1 - \sigma(s - \mu) \to 0$, we observe that the update signal vanishes, reflecting that learning pressure reduces once margin is satisfied.

This analysis confirms that the model primarily updates its parameters to increase the score difference $s$ when $s \leq \mu$. Once $s > \mu$, the incentive to further increase $s$ diminishes. Therefore, a larger $\mu$ maintains learning pressure across a wider range of $s$ values, promoting more substantial separation between preferred and non-preferred responses.

\section{RM Training Details}\label{app:train_de}
Without specific statement, we set $\gamma=0.5$ in \eqref{eq:compute_margin} and $\beta=0.1$ in OT. During reward model training, we adopt the LoRA and train all the reward models for 2 epochs using a learning rage of $4\times10^{-5}$. The inputs are truncated by 1024 tokens and more detailed hyper-parameters can be found in Table~\ref{hyper-params}. 
\begin{table}[h]
\centering
\caption{\small{Implementation details.}}
\resizebox{0.6\textwidth}{!}{
\begin{tabular}{c|c}
\toprule
Quantization & bf16 \\
LoRA $r$ & 32 \\
LoRA $\alpha$ & 64 \\
LoRA dropout & 0.05 \\
Optimizer & Adamw\_hf \\
Global Batch Size & 128 \\
Learning Rate & $2\times10^{-6}$ \\
Learning Rate Scheduler & cosine \\
RM-GPUs & $1\times4$ GPU Cards \\
RLHF-GPUs & $1\times8$ GPU Cards \\
Warmup Ratio & 0.03 \\ \bottomrule
\end{tabular}
}\label{hyper-params}
\end{table}

\section{PPO Training and Evaluation Details}\label{app:ppo_train_eval_de}
For our PPO experiment, we fine-tune two distinct models using 20,000 samples from the alpaca-gpt4-data-en dataset~\citep{peng2023gpt4llm}. The first model, Llama3.1-8B-Instruct\footnote{\url{https://huggingface.co/meta-llama/Llama-3.1-8B-Instruct}}, has undergone post-training that includes both DPO and RLHF. The second, OpenRLHF-Llama3-8B-SFT\footnote{\url{https://huggingface.co/OpenRLHF/Llama-3-8b-sft-mixture}}, is an instruction-following version built upon Llama3-8B-Base, without RLHF post-training stage. We conduct the PPO training using the ms-swift framework\footnote{\url{https://github.com/modelscope/ms-swift}} with its default training configuration. All benchmark evaluations are subsequently performed using the ms-evalscope framework\footnote{\url{https://github.com/modelscope/evalscope}}. Our evaluation protocol utlize few-shot settings for GSM8K (4-shot)~\citep{cobbe2021trainingverifierssolvemath}, Race (3-shot)~\citep{lai2017racelargescalereadingcomprehension}, and TriviaQA (5-shot)~\citep{joshi2017triviaqalargescaledistantly}, while all other benchmarks (i.e., Hellaswag~\citep{zellers2019hellaswagmachinereallyfinish}, IFeval~\citep{zhou2023instructionfollowingevaluationlargelanguage}, MMLU~\citep{hendrycks2021measuringmassivemultitasklanguage}, ProcessBench~\citep{zheng2025processbenchidentifyingprocesserrors}, BBH~\citep{suzgun2022challengingbigbenchtaskschainofthought}, and Humaneval~\citep{chen2021evaluatinglargelanguagemodels}) are assessed in a zero-shot setting. We report accuracy as the primary metric for all tasks, with the exception of Humaneval, for which we report the Pass@1 score.

\end{document}